\documentclass[10pt,twocolumn,letterpaper]{article}

\usepackage{cvpr}
\usepackage{times}
\usepackage{epsfig}
\usepackage{graphicx}
\graphicspath{{images//}}
\usepackage{amsmath}
\usepackage{amssymb}
\usepackage{gensymb}

\usepackage[pagebackref=true,breaklinks=true,letterpaper=true,colorlinks,bookmarks=false]{hyperref}

\usepackage[square,numbers,sort&compress]{natbib}
\usepackage[normalem]{ulem}

\cvprfinalcopy 

\ifcvprfinal\fi

\begin{document}

\title{Crowd Counting by Adapting Convolutional Neural Networks with Side Information}

\author{Di Kang\\
City University of Hong Kong\\
{\tt\small dkang5-c@my.cityu.edu.hk}
\and
Debarun Dhar\\
City University of Hong Kong\\
{\tt\small ddhar2@cityu.edu.hk}
\and
Antoni B. Chan\\
City University of Hong Kong\\
{\tt\small abchan@cityu.edu.hk}
}

\maketitle

\begin{abstract}

Computer vision tasks often have side information available that is  helpful to solve the task.
For example, for crowd counting, the camera perspective (e.g., camera angle and height) gives a clue about the appearance and scale of people in the scene.
While side information has been shown to be useful for counting systems using traditional hand-crafted features, it has not been fully utilized in counting systems based on deep learning.
In order to incorporate the available side information, we propose an adaptive convolutional neural network (ACNN), where the convolutional filter weights adapt to the current scene context via the side information.
In particular, we model the filter weights as a low-dimensional manifold, parametrized by the side information, within the high-dimensional space of filter weights.
With the help of side information and adaptive weights, the ACNN can disentangle the variations related to the side information, and extract discriminative features related to the current context.
Since existing crowd counting datasets do not contain ground-truth side information, we collect a new dataset with the ground-truth camera angle and height as the side information.
On experiments in crowd counting, the ACNN improves counting accuracy compared to a plain CNN with a similar number of parameters.
We also apply ACNN to image deconvolution to show its potential effectiveness on other computer vision applications.

\end{abstract}

\section{Introduction}

Automatic analysis of crowded scenes in images and videos has important real-world applications in surveillance, crowd management and traffic control. Estimating the number of people is one important task for all these potential applications. Crowd counting is still a very challenging task due to complicated backgrounds, severe occlusion, low-resolution imaging, perspective distortion, and different appearances caused by different camera tilt angles.
\par
Early counting by detection~\cite{part_detector_ICCV05} and counting by tracking~\cite{brostow2006unsupervised,KLT_tracker_CVPR2006} methods have gradually been replaced by recent regression based methods~\cite{Chan2008,Ryan2009,NIPS2010_4043,Arteta2014,Fiaschi2012,Zhang2015,Zhang2015c}, and especially  crowd density estimation methods~\cite{NIPS2010_4043,Arteta2014,Fiaschi2012,Zhang2015,Zhang2015c}.
Regression-based methods bypass the need to explicitly detect every person, and their performance does not decrease dramatically when the scene is very crowded.
With crowd density estimation, each pixel in the crowd density map represents the fraction of people in that location, and the crowd count is obtained by integrating over a region in the density map.
Most recent state-of-the-art counting systems are based on using convolutional neural networks (CNN) for crowd density estimation~\cite{Zhang2015,Zhang2015c} because of the powerful feature extraction and learning capabilities of CNNs.

\par
\begin{figure}[tb]
\centering
\includegraphics[width=0.98\linewidth]{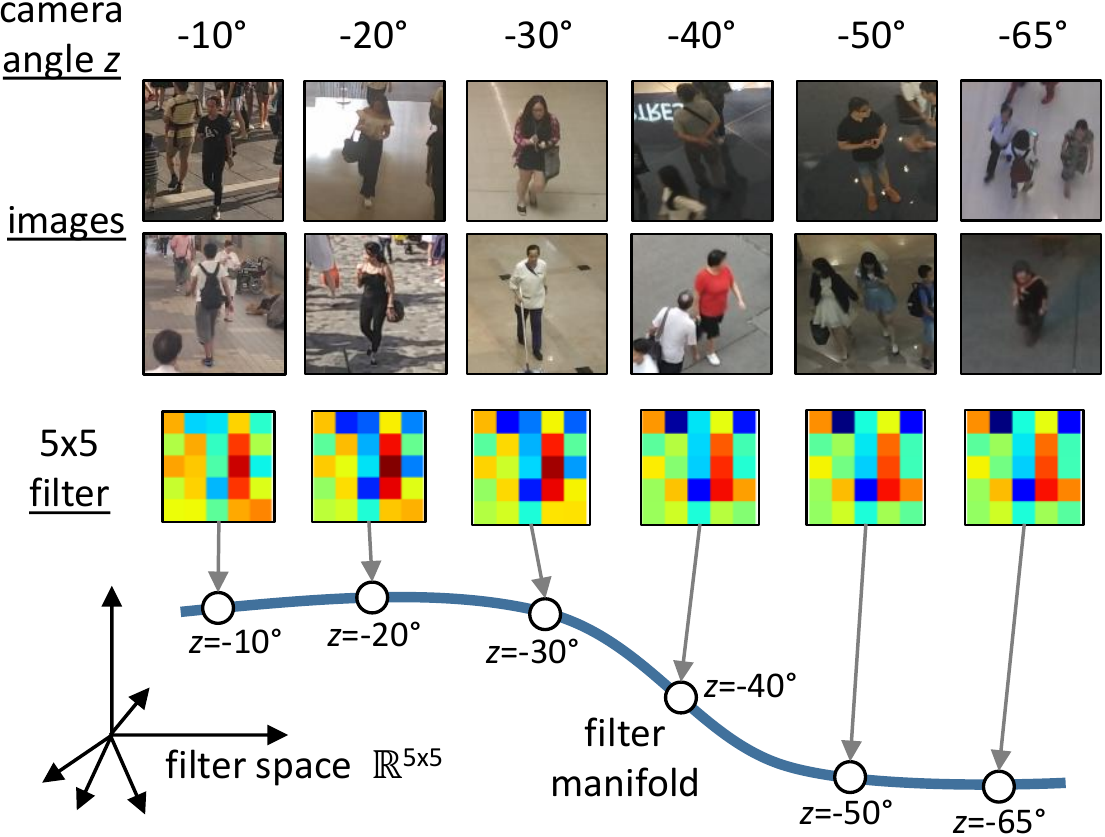}
\caption{(top) changes in people's appearance due to camera angle, and the corresponding changes in a convolution filter;
(bottom) the filter manifold parametrized by the camera angle. Best viewed in color.
}
\label{fig:manifold}
\end{figure}

Previous works have also shown that using side information about the scene, e.g., the scene perspective,  can help to improve crowd counting accuracy \cite{Chan2008,Chan2009,Chan2012}. In particular, when extracting hand-crafted features (e.g., edge and texture statistics) \cite{Chan2008,Chan2012,Chan2009,NIPS2010_4043,Arteta2014,Idrees2013} use scene perspective normalization, where a ``perspective weight'' is applied at each pixel location during feature extraction, to adjust for the scale of the object at that location (see Fig.~\ref{fig:image-perspective-density}b).
To overcome the difficulty caused by perspective distortion, some CNN methods \cite{Zhang2015} normalize the input patch size according to perspective value, which mainly adjusts for the scale of the person.
However, this also discards important information that could be helpful in analyzing an image patch.
For example, people have different appearances when viewed from different camera angles
(see Fig.~\ref{fig:manifold} top). Even within the same scene, a person's appearance changes  as they move along the ground-plane, due to changes in the relative angles to the camera sensor.
\cite{Zhang2015c} uses a multi-column structure where each column has a different receptive field to encourage different columns to capture features of different scale, but without explicit supervision.
Hence, in the current CNN methodologies, the existing side information has not been explicitly incorporated into the architecture.
As a result, all the difficulties due to various contexts, including different backgrounds, occlusion, perspective distortion and different appearances caused by different camera tilt angles are entangled, which may introduce an extra burden on the CNNs during training.

\par
Our solution in this paper is to disentangle the context variations explicitly in the CNN. We propose a CNN architecture, called adaptive CNN (ACNN), that uses side information (e.g., the perspective weight) as an auxiliary input to adapt the CNN to different scene contexts (e.g., appearance changes from high/low angle perspectives, and scale changes due to distance) within or across different scenes.
Specifically, we consider the filter weights in each convolutional layer as points on a low-dimensional manifold, which is parametrized by the side information and generated by a sub-network.
The ``filter manifold'' is estimated during training, resulting in different convolution filters for each scene context, which disentangles the context variations related to the side information (auxiliary input).
In the ACNN, the convolutional layers focus only on those features most suitable for the current context defined by the side information, as compared to traditional CNNs that use a fixed set of filters over all contexts. In other words, the feature extractors are tuned for each context.
Furthermore,
ACNN using the camera parameters (camera tilt angle and height) as the side information can perform cross-scene counting, without any further fine-tuning stage, which is more suitable for quick deployment.
Note that the camera tilt angle and height can be easily obtained using the accelerometer tilt sensor of a smartphone and a laser range finder.

\par
In order to test the performance of our ACNN with camera tilt angle and height on cross-scene counting, we collect a new dataset covering a wide range of angles (-10\degree\ to -65\degree) and heights (2.2m to 16.0m), containing different appearances of people from different viewpoints.
Besides the crowd counting application, side information is often available for many computer vision tasks, but has been mostly ignored by current CNN architectures. For example, in the image de-blurring task, the blur kernel parameters are sometimes known or can be estimated. Using these kernel parameters as side-information, a single ACNN can be trained to deblur images for any setting of the kernel parameters. In contrast, using a standard CNN would require training a separate CNN for each combination of kernel parameters, which is costly if the set of parameter combinations is large.

\par
The contributions of this paper are 4-fold:
1) We propose a method to incorporate the side information directly into CNN by using an adaptive convolutional layer whose weights are generated according to available side information, and validate it on various crowd counting datasets and a non-blind image deconvolution task;
2) Using ACNN, we achieve state-of-the-art or comparable to state-of-the-art performance on two crowd counting datasets;
3) We collect a new crowd counting dataset covering a wide range of viewpoints and its corresponding side information, i.e. camera tilt angle and camera height;
4) we show that it is possible to train one network for image deconvolution with different kernel parameters using ACNN and the extra adaptability introduced by ACNN increases its performance compared with a plain CNN.

\section{Related work}

\subsection{Adapting neural networks}

The performance of a CNN will be affected if the test set is not from the same data distribution as the training set~\cite{Zhang2015}.
A typical approach to adapting a CNN to new data is to select a base pre-trained CNN model,
e.g. AlexNet~\cite{AlexNet}, VGG-net~\cite{Simonyan14c}, or ResNet~\cite{He2015} trained on ImageNet,
and then fine-tune the model weights to make them more suitable for the specific task.
\cite{Zhang2015} adopts a similar strategy -- train the model on the whole dataset and then fine-tune
using a subset of image patches that are similar to the test scene.

\par
Another approach is to adapt the input data cube so that the extracted features and the subsequent classifier/regressor are better matched.
\cite{Kavukcuoglu2015} proposes a trainable ``Spatial Transformer'' unit that applies an image transformation (warping)
to register the input image to a standard form before the convolution layer. The functional form of the image transformation must be known, and the transformation parameters are estimated from the image.
Because it operates directly on the image, \cite{Kavukcuoglu2015} is limited to 2D image transformations, which work well for 2D planar surfaces in an image (e.g., text on a flat surface), but cannot handle viewpoint changes of 3D objects (e.g. people).
In contrast, our ACNN changes the feature extraction layers to match the 3D viewpoint, and does not require the geometric transformation to be known.

\par
Most related to our work are dynamic convolution~\cite{Klein2015} and dynamic filter networks~\cite{DeBrabandere2016}, which use the input image to dynamically generate the filter weights for convolution. However, their purpose for dynamically generating filters is quite different from ours. \cite{Klein2015,DeBrabandere2016} focus on image prediction tasks (e.g., predicting the next frame from the previous frames), and the dynamically-generated filters are mainly used to transfer a pixel value in the input image to a new position in the output image (e.g., predicting the movement of pixels between frames).
These input-specific filters are suitable for low-level tasks, i.e. the input and the output are both in the same space (e.g., images).
But for high-level tasks, dramatically changing features with respect to its input is not helpful for the  end-goal of classification or regression.
In contrast, our purpose is to include domain knowledge/side information into supervised learning (regression), by learning how the discriminative image features and corresponding filters change with respect to the side information (i.e., the filter manifold). Hence, in our ACNN, the filter weights are generated from an auxiliary input corresponding to the side information.

\par
Finally, one advantage of \cite{Kavukcuoglu2015,Klein2015,DeBrabandere2016} is that no extra information or label is needed. However, this also means they cannot effectively utilize the available side information, which is common in various computer vision tasks and has been shown to be helpful for traditional hand-crafted features~\cite{Chan2008}.

\subsection{Crowd counting}

Early works in crowd counting~\cite{part_detector_ICCV05,brostow2006unsupervised,KLT_tracker_CVPR2006} explicitly detect and/or track every object in the scene. But as the scene becomes more crowded, the performance of these detection- and tracking-based methods decreases dramatically.
Counting by regression bypasses the difficulty of the individual object detection problem, and works well
when the scene is very crowded or in low resolution. Initial works directly regress from either global features or local features directly to image count \cite{Cho1999,Chen2012,Chan2008,Chan2012,Idrees2013,Chen2013} or blob count~\cite{Kong2006,Ryan2009}, discarding almost all the spatial information within the region, and hence can only be used for counting.
\cite{NIPS2010_4043} proposes the concept of an object density map whose integral over any region equals to the number of objects in that region, achieving state-of-the-art counting performance. The spatial distribution of the objects is preserved in the density map, which also makes it useful for detection \cite{Ma_2015_CVPR} and tracking \cite{Rodriguez2011}.
Most of the recent object counting algorithms adopt the density estimation approach~\cite{Arteta2014,Fiaschi2012,Zhang2015,Zhang2015c} and achieve the state-of-the-art performance.

\par
The two CNN-based methods~\cite{Zhang2015,Zhang2015c} show strong cross-scene prediction capability, due to the learning capacity of deep neural networks.
Specifically, \cite{Zhang2015c} uses a multi-column CNN with different receptive field sizes in order to encourage different columns to capture features at different scales (without explicit supervision), while
\cite{Zhang2015} introduces an extra fine-tuning stage so that the network can be adapted to a new scene.
In contrast to \cite{Zhang2015,Zhang2015c}, we propose to use the existing side information (e.g. perspective weight) to adapt the convolutional layers to different contexts. With the adaptive convolutional layers, only the discriminative features suitable for the current context are extracted. Our experiments show that moving parameters from static layers to adaptive layers yields stronger learning capability and adaptability.

\subsection{Image deconvolution}
Existing works~\cite{Eigen2013,Xu2014,Burger2012} demonstrate that CNNs can be used for image deconvolution and restoration. With non-blind deblurring, the blur kernel is known and the goal is to recover the original sharp image.
Inspired by traditional optimization-based deconvolution methods, \cite{Xu2014} proposes to use a deep CNN to solve the deconvolution problem and the following artifact removal problem by concatenating a deep deconvolution CNN and a denoising CNN.
Specifically, \cite{Xu2014} requires a separate network to be trained for each blur kernel family and kernel parameter.
\cite{Burger2012} trains a multi-layer perceptron (MLP) to denoise images corrupted by additive white Gaussian (AWG) noise of different levels. They incorporate the side information (the standard deviation of AWG) by simply concatenating it to the vectorized image patch input.
In this paper, we use the kernel parameters as the auxiliary input, and apply the ACNN to train a single network for a blur kernel family (for all its parameter values), rather than for each possible kernel parameter. During prediction, the kernel parameter (auxiliary input) generates the appropriate deblurring filters, without the need for additional training.
In summary, there are two main advantages of using ACNN for deconvolution:
1) only one network is needed for each blur kernel family, which is especially useful for those kernels where there are too many parameters to enumerate;
2) ACNN provides more flexibility and performs better for both the kernel parameters used for training, as well as, interpolating between the training set.

\section{Adaptive CNN}

In this section, we introduce the adaptive convolutional layer and the ACNN for crowd density estimation.

\subsection{Adaptive convolutional layer}

Consider a dataset containing different viewpoints of people, and we train a separate CNN to predict the density map for each viewpoint. For two similar viewpoints, we expect that the two trained CNNs have similar convolutional filter weights, as a person's appearance varies gradually with the viewpoint (see~\ref{fig:manifold} top).  Hence, as the viewpoint changes smoothly, the convolutional filters weights also change smoothly, and thus sweep a low-dimensional manifold within the high-dimensional space of filter weights (see~\ref{fig:manifold} bottom).

\begin{figure}[htb]
\centering
\includegraphics[width=0.88\linewidth]{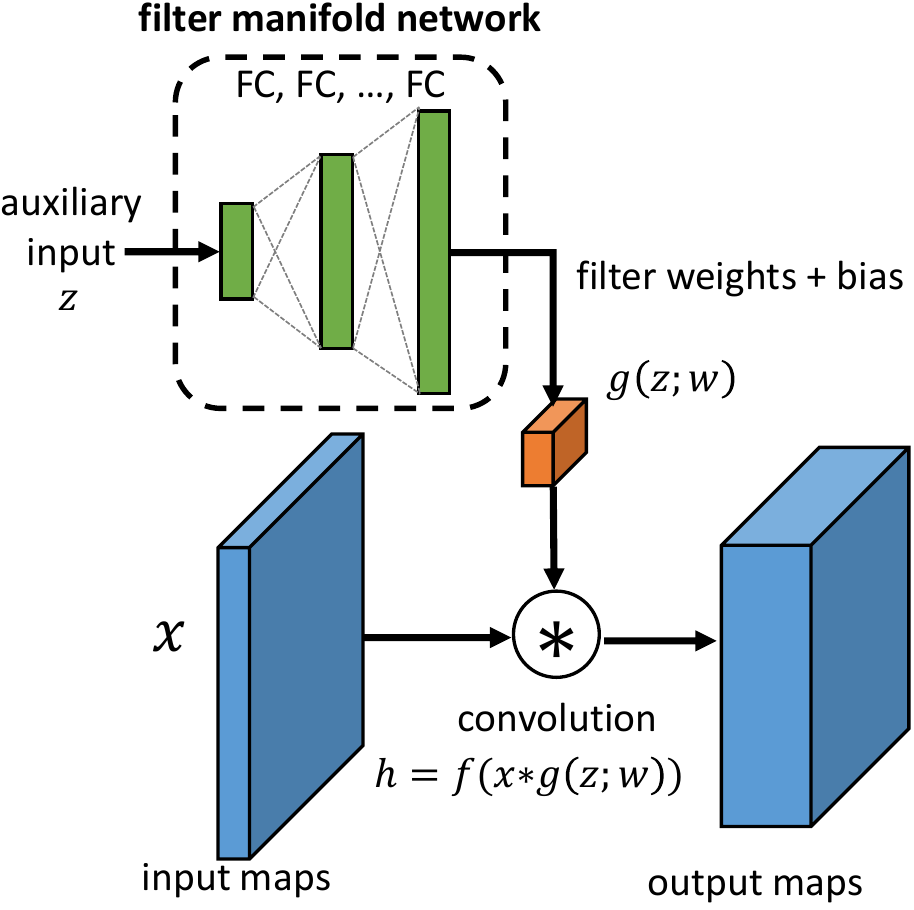}
\caption{The proposed adaptive convolution layer with filter manifold network (FMN). The auxiliary input generate the filter weights, which are then convolved with the input maps.
}
\label{fig:adaptive_conv}
\end{figure}

Following this idea, we use an {\em adaptive convolutional layer}, where the convolution filter weights are the outputs of a separate ``filter-manifold'' network (FMN, see Fig.~\ref{fig:adaptive_conv}).
In the FMN, the auxiliary input (side information, such as camera tilt angle) is fed into a set of fully-connected layers with increasing dimension (a multi-layer perceptron, similar to the decoder stage of an auto-encoder) with the final layer outputting the convolution filter weights. The output of the FMN is reshaped into a 4D tensor of convolutional filter weights (and bias), and convolved with the input image.
Note that in contrast to the traditional convolutional layer, whose filter weights are fixed during the inference stage, the filter weights of an adaptive convolutional layer change with respect to the auxiliary input.

\par
Formally, the adaptive convolutional layer is given by
\begin{align}
  h = f( x \ast g(z; w) ),
\end{align}
where $z$ is the auxiliary input, $g(\cdot; w)$ is the filter manifold network with tunable weights $w$, $x$ is the input image, and $f(\cdot)$ is the activation function.\footnote{To reduce clutter, here we do not show the bias term for the convolution.}
Training the adaptive convolutional layer involves updating the FMN weights $w$, thus learning the filter manifold as a function of the auxiliary input.
During inference, the FMN interpolates along the filter manifold using the auxiliary input, thus adapting the filter weights of the convolutional layer to the current context.
Hence adaptation does not require fine-tuning or transfer learning.
The adaptive convolution layer can disentangle different input variations with respect to the side information, e.g., variations in appearance due to changes in viewpoint, by using different filter weights for each context.

\subsection{Adaptive CNN for crowd counting}

\begin{figure}[tb]
\centering
\begin{tabular}{c}
  \includegraphics[width=0.98\linewidth]{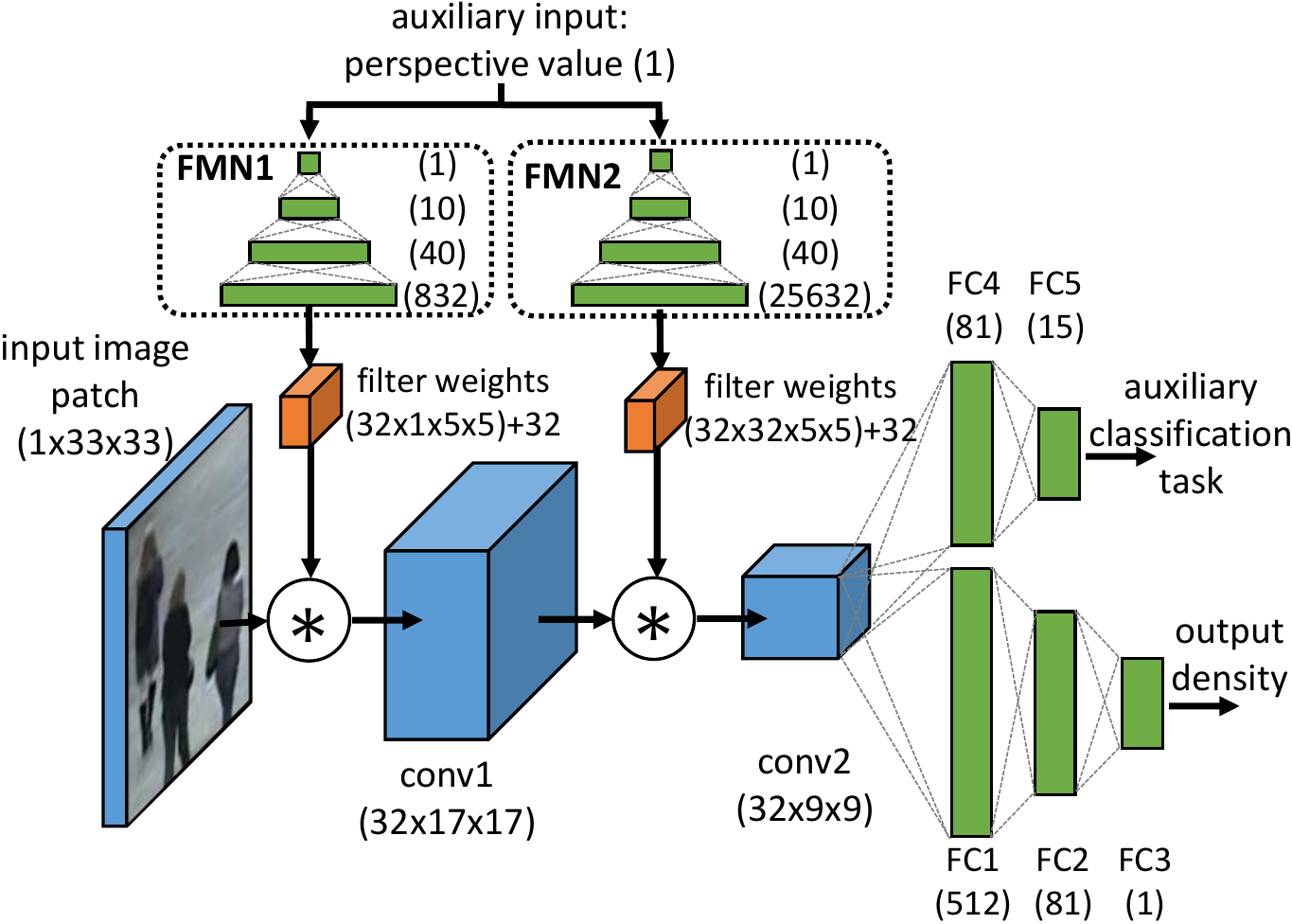} \\
\end{tabular}
\caption{The architecture of our ACNN with adaptive convolutional layers for crowd density estimation.}
\label{fig:arch}
\end{figure}

Density map estimation is not as high-level a task as recognition. Since the upper convolutional layers extract more abstract features, which are probably not very helpful according to both traditional methods~\cite{Chan2008,NIPS2010_4043} and deep CNN methods~\cite{Zhang2015,Zhang2015c}, we will not use too many convolutional layers, taking both accuracy and speed into consideration. Generally, two or three convolutional layers are used for the convolutional stage.
Fig.~\ref{fig:arch} shows our ACNN for density map estimation.
The input is an image patch, while the output is the crowd density at the center of the patch.
All the convolutional layers use the ReLU activation, and each convolutional layer is followed by a local response normalization (LRN)~\cite{AlexNet} layer and a max pooling layer.
The auxiliary input for the FMN is the perspective value for the image patch in the scene, or the camera tilt angle and camera height.
For the fully-connected stage, we use multi-task learning to improve the performance and robustness of the learned features~\cite{Li2015,Zhang2014,Sun2014,Zhang2015}.
In particular, the main regression task predicts the crowd density value, while an auxiliary classification task classifies the image patch according to the number of people within.

\par
In the ACNN,  the extracted features of the convolutional layers are adapted to the current scene context, as compared to a traditional CNN that uses a fixed set of filters over all scene contexts.
For CNNs with fixed filters, the filter response patterns change with  viewpoint, and these complex variations must be summarized in the subsequent fully-connected layers.
In contrast, for ACNN, the filters themselves change with the viewpoint resulting in filter responses that are consistent across viewpoints.

\par
Note that the adaptive convolutional layer has more parameters than a standard convolutional layer with the same number of filters and the same filter spatial size -- the extra parameters are in the hidden layers of the FMN. However, since the filters themselves adapt to the scene context, an ACNN can be effective with fewer feature channels,
and the parameter savings can be moved to the FMN (e.g. see Table~\ref{tab:ACNNparams}).
Hence, if side information is available, a standard CNN can be converted into an ACNN with a similar number of parameters, but with better learning capability. We verify this property in the experiments.

\begin{table}[tbh]
\centering
\small
\begin{tabular}{cr@{\hspace{0.08cm}}lr@{\hspace{0.08cm}}l}
  \hline
  Layer & CNN &         		&  ACNN & \\
  \hline
  FMN1  & -- &          		& 34,572 &(832) \\
  conv1 & 1,664 &(64)   		& 0 &(32) \\
  FMN2  & -- &         		& 1,051,372 &(25,632) \\
  conv2 & 102,464 &(64) 		& 0 &(32) \\
  \hline
  FC1   & 2,654,720 &(512) 	& 1,327,616 &(512) \\
  FC2   & 41,553 &(81) 		& 41,553 &(81) \\
  FC3   & 82 &(1) 			& 82& (1) \\
  \hline
  FC4   & 419,985 &(81) 		& 210,033 &(81) \\
  FC5   & 1,312 &(15) 		& 1,312& (15)
  \\
  \hline
  total & 3,221,780 &		& 2,666,540 & \\
  \hline
\end{tabular}
\caption{Comparison of number of parameters in each layer of the ACNN in Fig.~\ref{fig:arch} and an equivalent CNN. The number in parenthesis is the number of convolutional filters, or the number of outputs of the FMN/fully-connected (FC) layer.}
\label{tab:ACNNparams}
\end{table}

\section{New crowd counting dataset with extrinsic camera parameters}

Camera extrinsic parameters, such as camera tilt angle and camera height, are useful types of side information for crowd counting.
The camera tilt angle affects a person's appearance (see Fig.~\ref{fig:manifold} top), while the height affects the scale.
Note that the camera tilt angle/height can be used to estimate the perspective map, whose values indicate the size of a person appearing at each location.
As existing datasets do not contain the side information of extrinsic camera parameters, here we collect a new dataset consisting of indoor/outdoor scenes from various camera angles and heights.
The scenes were captured using a smartphone camera, which was placed on a tripod to keep it stable. The camera tilt angle was recorded using the accelerometer tilt-sensor of the smartphone, and the height of the camera to the ground-plane was  measured using a laser range finder.
The perspective map for each scene is estimated from the camera extrinsic parameters using the method described in Section~\ref{ss:perspective_est}.

\par
\begin{figure}[htb]
\centering
\small
\begin{tabular}{@{}c@{\hspace{1.5mm}}c@{\hspace{1.5mm}}c@{}}
  (a) image & (b) perspective map & (c) GT density map \cite{NIPS2010_4043} \\
  \includegraphics[width=0.29\linewidth]{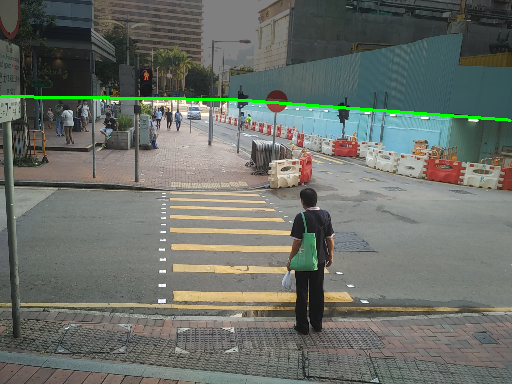} &
  \includegraphics[width=0.32\linewidth]{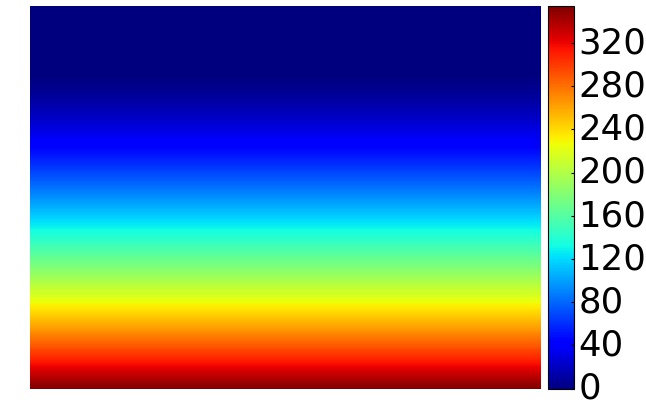} &
  \includegraphics[width=0.29\linewidth]{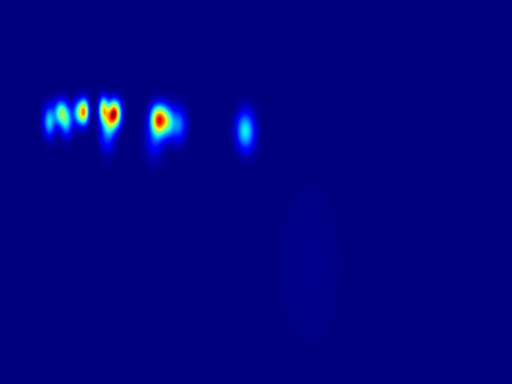} \\
  \includegraphics[width=0.29\linewidth]{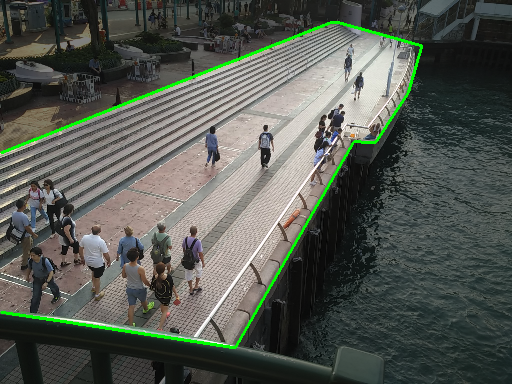} &
  \includegraphics[width=0.32\linewidth]{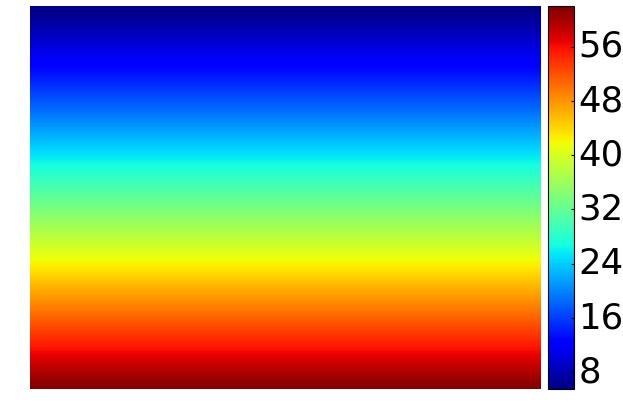} &
  \includegraphics[width=0.29\linewidth]{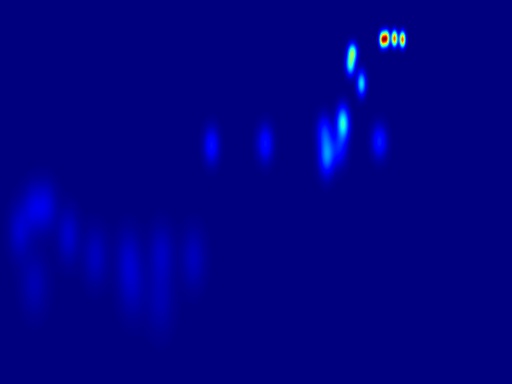} \\
  \includegraphics[width=0.29\linewidth]{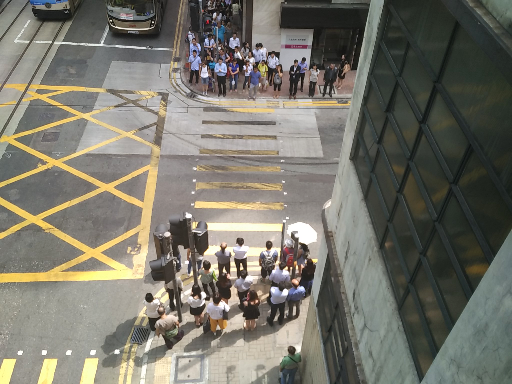} &
  \includegraphics[width=0.32\linewidth]{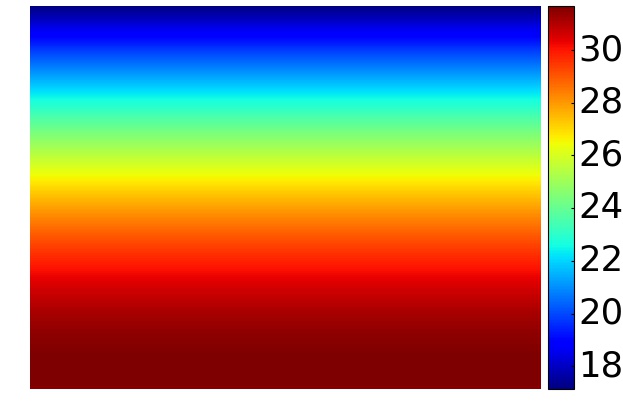} &
  \includegraphics[width=0.29\linewidth]{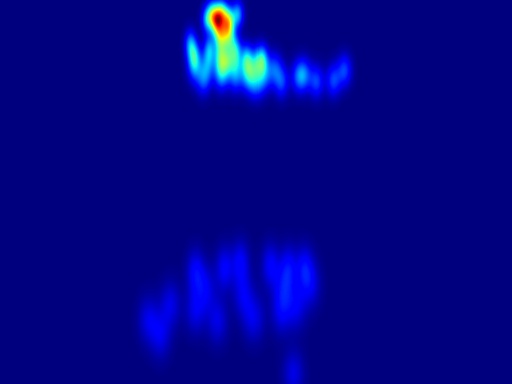} \\
  \includegraphics[width=0.29\linewidth]{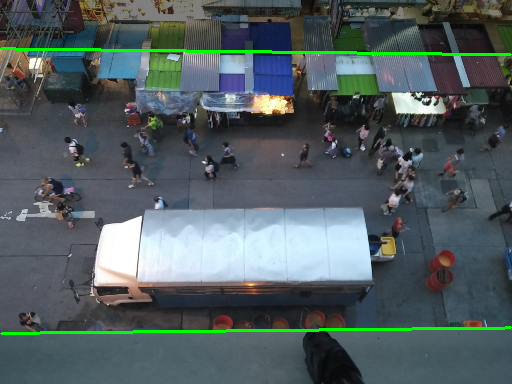} &
  \includegraphics[width=0.32\linewidth]{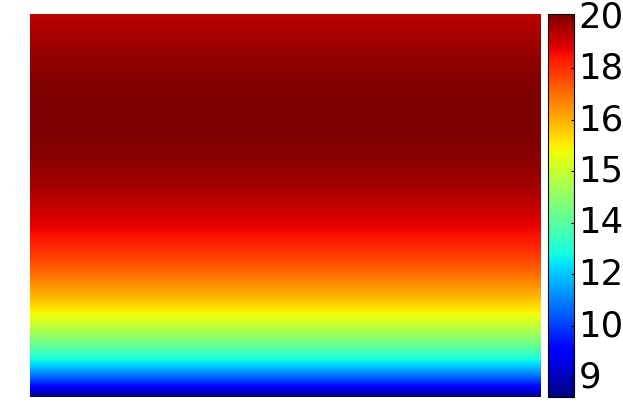} &
  \includegraphics[width=0.29\linewidth]{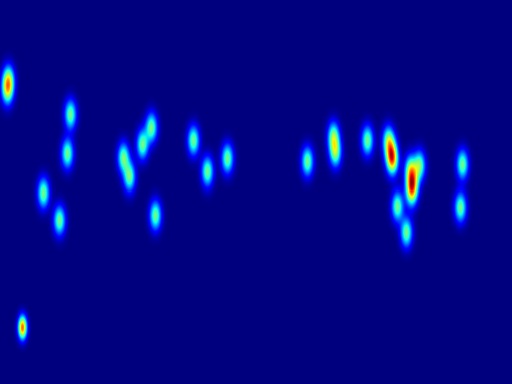} \\
\end{tabular}
\caption{Example images along with their estimated perspective maps, and ground-truth density maps from different scenes in our dataset}
\label{fig:image-perspective-density}
\end{figure}
\begin{figure}[htb]
\centering
\begin{tabular}{c}
  \includegraphics[width=0.98\linewidth]{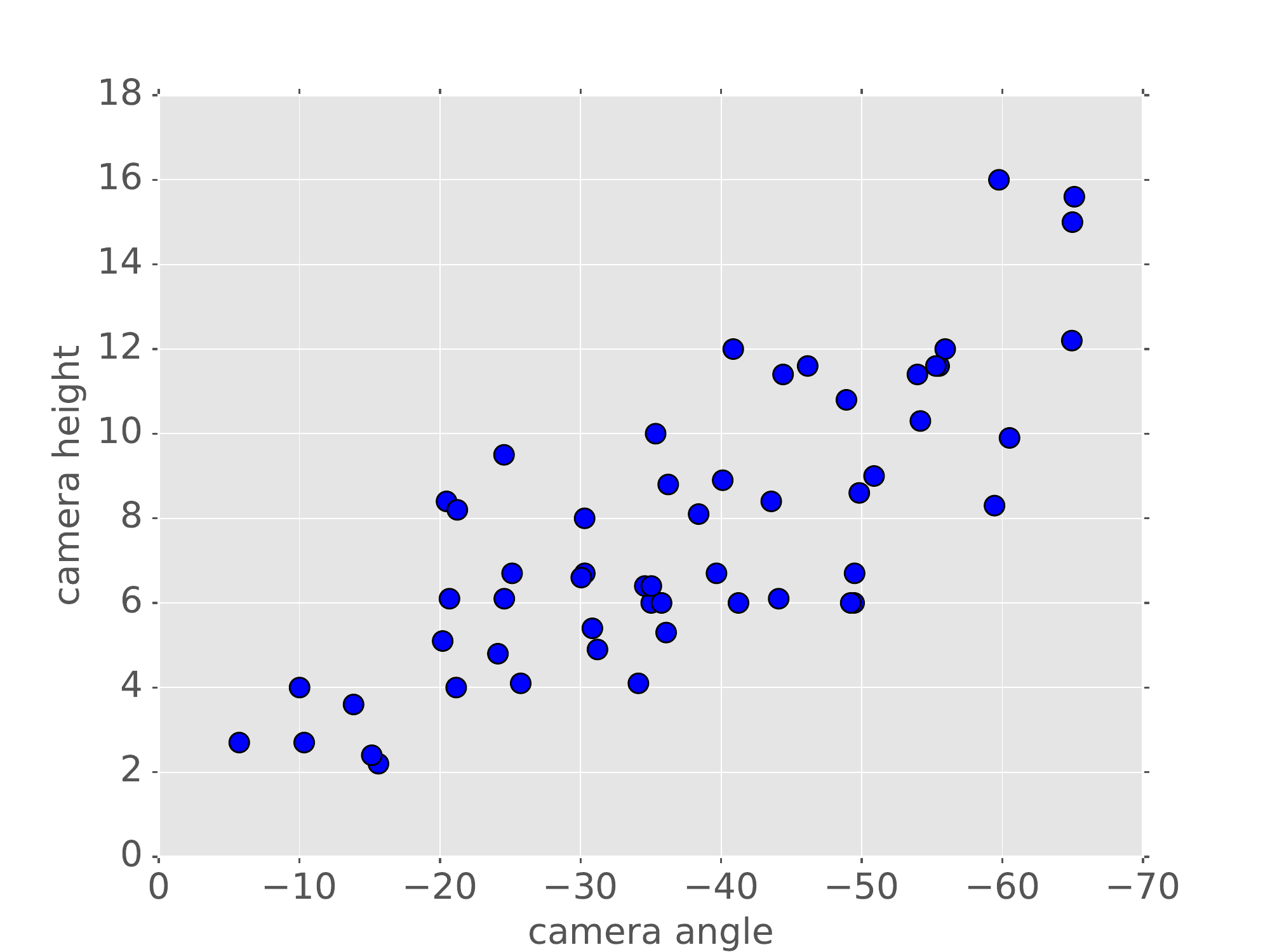} \\
\end{tabular}
\caption{Distribution of camera tilt angle and height for scenes in our new dataset.}
\label{fig:dataset_distribution}
\end{figure}

Fig.~\ref{fig:image-perspective-density} shows several example scenes, while Fig.~\ref{fig:dataset_distribution} shows the distribution of camera tilt angle and height for the scenes.
The new dataset contains 55 scenes, covering a camera tilt angle range of [-10\degree, -65\degree] and a height range of [2.2, 16.0] meters.
Each scene contains 58 images on average, with a time delay of $11\pm20$ seconds between images. The total number of images is 3,191, collected over $\sim$10 hours.
The training set consists of 43 scenes (2,503 images; 78,592 people), and the test set comprises 12 scenes (688 images; 28,191 people).
Similar to~\cite{Zhang2015,Zhang2015c}, the ground truth density map used for training and testing is generated using a Gaussian density with standard deviation that varies according to the perspective map ($ \sigma_h=\frac{1}{5}M_p, \sigma_v=\frac{1}{2}M_p$), where $M_p$ is the perspective value at location $p$.
However, instead of using two Gaussians to form a human shape (head and body), we use only a single elliptical Gaussian (for  the body), since in our dataset the appearance of people changes significantly due to a larger range of camera tilt angles, compared to WorldExpo \cite{Zhang2015}.

\subsection{Perspective map estimation from camera parameters} \label{ss:perspective_est}

We use the camera tilt angle, camera height and the camera's vertical field of view (FOV) to estimate the perspective map of a scene without manual labeling (c.f., \cite{Zhang2015}).
Following the convention in~\cite{Zhang2015}, the value in the perspective map is directly proportional to the size of a person appearing at the corresponding location in the image.

\par
\begin{figure}[tb]
\centering
\includegraphics[width=0.98\linewidth]{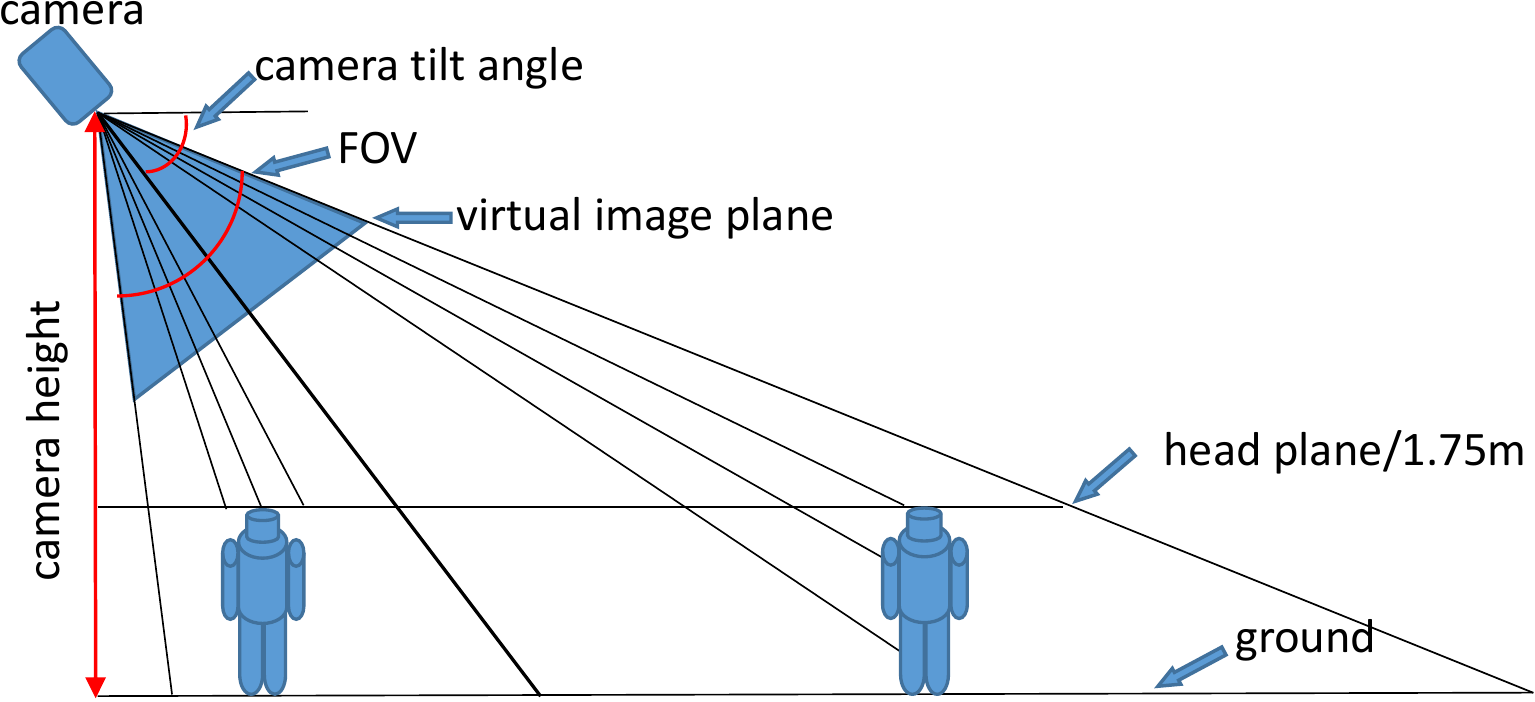}
\caption{Geometric relationships used to estimate the perspective map for a scene.
}
\label{fig:perspective_est}
\end{figure}

Since we know the camera's FOV and the image resolution, we can project a line through each image pixel to the scene (see Fig.~\ref{fig:perspective_est}).
The appearance of a person changes with the camera tilt angle: side-view for low camera tilt angles, and top-view for high camera tilt angles.
Using the scene geometry, we estimate the corresponding real-world projected depth $d$ and height $h$ for each pixel of the image, and obtain an estimate of a person's projected length on that image pixel as $a=\sqrt{d\cdot h}$, in order to represent both cases of a person's side-view and top-view.
The perspective value is $1/a$, which represents how many pixels are in 1 meter in the real world.
Refer to Fig.~\ref{fig:image-perspective-density} for the estimated perspective map, larger values indicate locations where people appear larger.

\section{Experiments}
We evaluate our proposed ACNNs on two tasks: crowd counting and image deconvolution.
For crowd counting, we use two crowd counting datasets: the newly collected dataset with camera tilt angle and camera height as side information, and the popular UCSD crowd counting dataset.
We also evaluate on image deconvolution to show the potential of ACNN for other computer vision tasks.
In order to make fair comparisons, we compare our ACNN with a standard CNN using traditional convolutional layers, but increase the number of filter channels in the CNN so that it has a similar total number of parameters as the ACNN.

\subsection{Crowd counting experiments}

\subsubsection{UCSD dataset}
Refer to Fig.~\ref{fig:arch} for the ACNN architecture used for the UCSD dataset. The image size is 238$\times$158, and a patch size of 33$\times$33 is used.
We test several variations of the ACNN:
v1) only the first convolutional layer is adaptive, with 64 filters for both of the convolutional layers;
v2) only the last convolutional layer is adaptive, with 64 filters for the first convolutional layer and 30 filters for its second/last convolutional layer;
v3) all the convolutional layers are adaptive, with 32 filters for all layers, which provides maximum adaptability.
The side information (auxiliary input) used for the FMN is the perspective value.
For comparison, we also test a plain CNN with a similar architecture but using standard convolution layers with 64 filters in each layer, and another plain CNN with input patch size normalization introduced in \cite{Zhang2015} (i.e., using larger patches for near-camera regions).
The count predictions in the region-of-interest (ROI) are evaluated using the mean absolute error (MAE) between the predicted count and the ground-truth.

\par
We first use the widely adopted protocol of ``max'' split, which uses 160 frames from 601:5:1400 for training, and the remaining parts (frames 1:600, 1401:2000) for testing.
The results are listed in Table~\ref{tab:ucsd_count}.
Our ACNN-v3, using two adaptive convolutional layers, offers maximum adaptability and has the lowest error (0.96 MAE), compared to the equivalent plain CNN and the reference methods.
In addition, using simple input patch size normalization  does not improve the performance as effectively as ACNN.
Examples of the learned filter manifolds are shown in Fig.~\ref{fig:UCSDfilters}.

\begin{table}[tb]
\centering
\small
\begin{tabular}{|c|c|}
  \hline
  Method                               & MAE   \\
  \hline
  MESA~\cite{NIPS2010_4043}            & 1.70  \\
  Regression forest~\cite{Fiaschi2012} & 1.70  \\
  RR~\cite{Arteta2014}                 & 1.24  \\
  \hline
  CNN-patch+RR~\cite{Zhang2015}        & 1.70  \\
  MCNN~\cite{Zhang2015c}               & 1.32  \\
  \hline
  CNN                            & 1.26 \\
  CNN (normalized patch)         & 1.26 \\
  ACNN-v1 (ours)	         & 1.23 \\
  ACNN-v2 (ours)	         & 1.14 \\
  ACNN-v3 (ours)                    & \bf{0.96} \\
  \hline
\end{tabular}
\caption{Comparison of mean absolute error (MAE) for counting with crowd density estimation methods  on the UCSD ``max'' split.}
\label{tab:ucsd_count}
\end{table}

\begin{figure}[tb]
\centering
\includegraphics[width=0.85\linewidth]{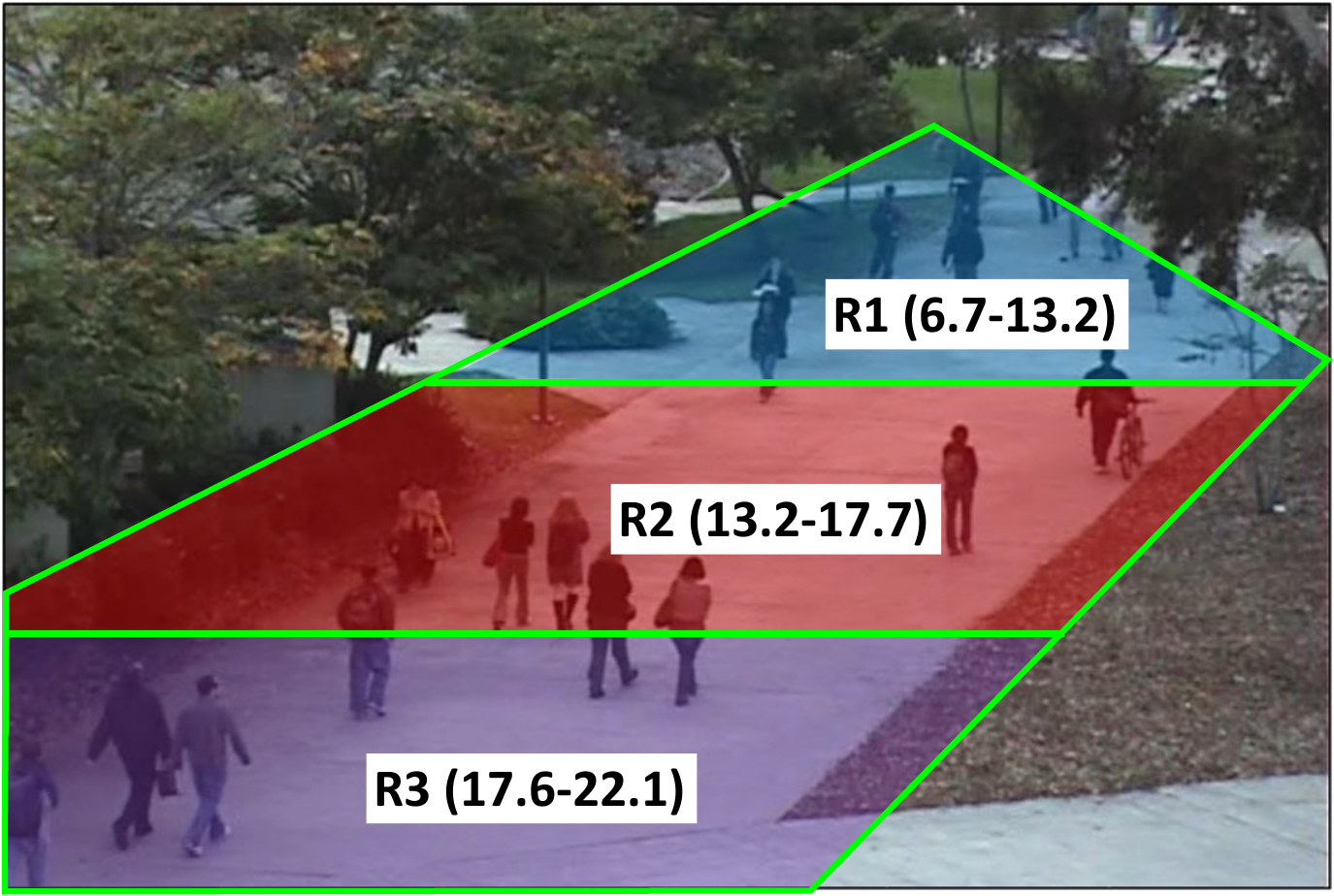}
\caption{UCSD dataset with 3 bar regions. The range of perspective values are shown in parentheses.}
\label{fig:ucsdbars-sm}
\end{figure}

In the next experiment we test the effect of the side information within the same scene. The ROI of UCSD is further divided into three bar regions of the same height (see Fig.~\ref{fig:ucsdbars-sm}).
The models are trained only on R1 and R3 on the training set, and tested on all three regions of the test set separately. The results are listed in Table~\ref{tab:ucsd_bar_count}.
After disentangling the variations due to perspective value, the performance on R1 has been significantly improved because the ACNN knows that it is different from the other regions.
Perspective values within R2 are completely unseen during training, but our ACNN still gives a comparable or slightly better performance than CNN.
This demonstrates that the filter manifold has learned a smoothly varying set of filters.

\begin{table}[tb]
\centering
\small
\begin{tabular}{|c|c|c|c|c|}
    \hline
    Method                     & R1  & R2 (unseen) & R3 & All \\
    \hline
    CNN                        & 1.83   & 1.06  & 0.62 & 1.17 \\
    ACNN-v1 (ours)             & 1.47   & 0.95  & 0.59 & 1.00 \\
    ACNN-v2 (ours)             & 1.22   & \bf{0.91}  & \bf{0.55} & \bf{0.89} \\
    ACNN-v3 (ours)             & \bf{1.15}   & 1.02  & 0.63 & 0.93 \\
    \hline
\end{tabular}
\caption{Comparison of MAE on three bar regions on the UCSD ``max'' split.}
\label{tab:ucsd_bar_count}
\end{table}

\newlength{\myw}
\setlength{\myw}{0.0625\linewidth}
\begin{figure}[tb]
\centering
\small
\begin{tabular}{@{}c@{}c@{}c@{}c@{}c@{}c@{}c@{}c@{}c@{}c@{}c@{}c@{}c@{}c@{}c@{}c@{}}
6.7 &$\cdots$&$\cdots$& 9.7 &$\cdots$&$\cdots$& 12.6 &$\cdots$&$\cdots$& 15.5 &$\cdots$&$\cdots$& 18.5 &$\cdots$&$\cdots$& 21.4
\\
\frame{\includegraphics[width=\myw]{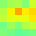}}&
\frame{\includegraphics[width=\myw]{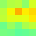}}&
\frame{\includegraphics[width=\myw]{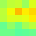}}&
\frame{\includegraphics[width=\myw]{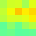}}&
\frame{\includegraphics[width=\myw]{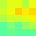}}&
\frame{\includegraphics[width=\myw]{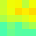}}&
\frame{\includegraphics[width=\myw]{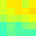}}&
\frame{\includegraphics[width=\myw]{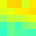}}&
\frame{\includegraphics[width=\myw]{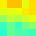}}&
\frame{\includegraphics[width=\myw]{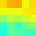}}&
\frame{\includegraphics[width=\myw]{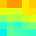}}&
\frame{\includegraphics[width=\myw]{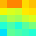}}&
\frame{\includegraphics[width=\myw]{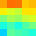}}&
\frame{\includegraphics[width=\myw]{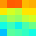}}&
\frame{\includegraphics[width=\myw]{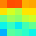}}&
\frame{\includegraphics[width=\myw]{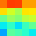}}
\\
\frame{\includegraphics[width=\myw]{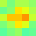}}&
\frame{\includegraphics[width=\myw]{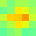}}&
\frame{\includegraphics[width=\myw]{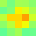}}&
\frame{\includegraphics[width=\myw]{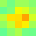}}&
\frame{\includegraphics[width=\myw]{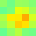}}&
\frame{\includegraphics[width=\myw]{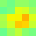}}&
\frame{\includegraphics[width=\myw]{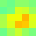}}&
\frame{\includegraphics[width=\myw]{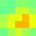}}&
\frame{\includegraphics[width=\myw]{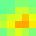}}&
\frame{\includegraphics[width=\myw]{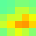}}&
\frame{\includegraphics[width=\myw]{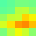}}&
\frame{\includegraphics[width=\myw]{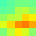}}&
\frame{\includegraphics[width=\myw]{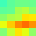}}&
\frame{\includegraphics[width=\myw]{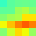}}&
\frame{\includegraphics[width=\myw]{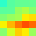}}&
\frame{\includegraphics[width=\myw]{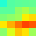}}
\\
\frame{\includegraphics[width=\myw]{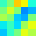}}&
\frame{\includegraphics[width=\myw]{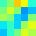}}&
\frame{\includegraphics[width=\myw]{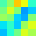}}&
\frame{\includegraphics[width=\myw]{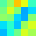}}&
\frame{\includegraphics[width=\myw]{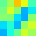}}&
\frame{\includegraphics[width=\myw]{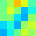}}&
\frame{\includegraphics[width=\myw]{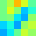}}&
\frame{\includegraphics[width=\myw]{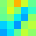}}&
\frame{\includegraphics[width=\myw]{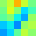}}&
\frame{\includegraphics[width=\myw]{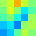}}&
\frame{\includegraphics[width=\myw]{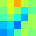}}&
\frame{\includegraphics[width=\myw]{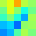}}&
\frame{\includegraphics[width=\myw]{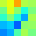}}&
\frame{\includegraphics[width=\myw]{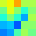}}&
\frame{\includegraphics[width=\myw]{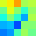}}&
\frame{\includegraphics[width=\myw]{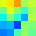}}
\end{tabular}
\caption{Examples of learned filter manifolds for the 2nd convolution layer. The numbers are the auxiliary input (perspective weight). Both the amplitude and patterns change, which shows the adaptability of the ACNN.}
\label{fig:UCSDfilters}
\end{figure}

\subsubsection{New dataset with extrinsic camera parameters}
The resolution of the new dataset is 512$\times$384, and 65$\times$65 patches are used.
The ACNN for this dataset contains three convolutional and pooling layers, resulting in the same output feature map size after the convolutional stage as in the ACNN for UCSD.
The three adaptive convolutional layers use 40, 40 and 32 filters of size 5-by-5 each. The side information (auxiliary inputs) are camera tilt angle and camera height (denoted as ``AH'').
The baseline plain CNN uses 64 filters of size 5-by-5 for all three convolutional layers.
Results for ACNN, the plain CNN, and MCNN are presented in Table~\ref{tab:ahgle_height_count}.
The plain CNN and MCNN \cite{Zhang2015c}, which do not use side information, obtain similar results.
Using side information (camera angle and height) with ACNN decreases the MAE, compared to the plain CNN.

\begin{table}[tb]
\centering
\small
\begin{tabular}{|c|c|}
  \hline
  Method                               	& MAE        \\
  \hline
  MCNN~\cite{Zhang2015c}               	& 8.80       \\
  CNN                            		& 8.72       \\
  ACNN-AH (ours)                        & 8.35       \\
  \hline
\end{tabular}
\caption{Counting results on our new dataset.}
\label{tab:ahgle_height_count}
\end{table}

\subsection{Image deconvolution}
\begin{figure}[tb]
\centering
\includegraphics[width=0.98\linewidth]{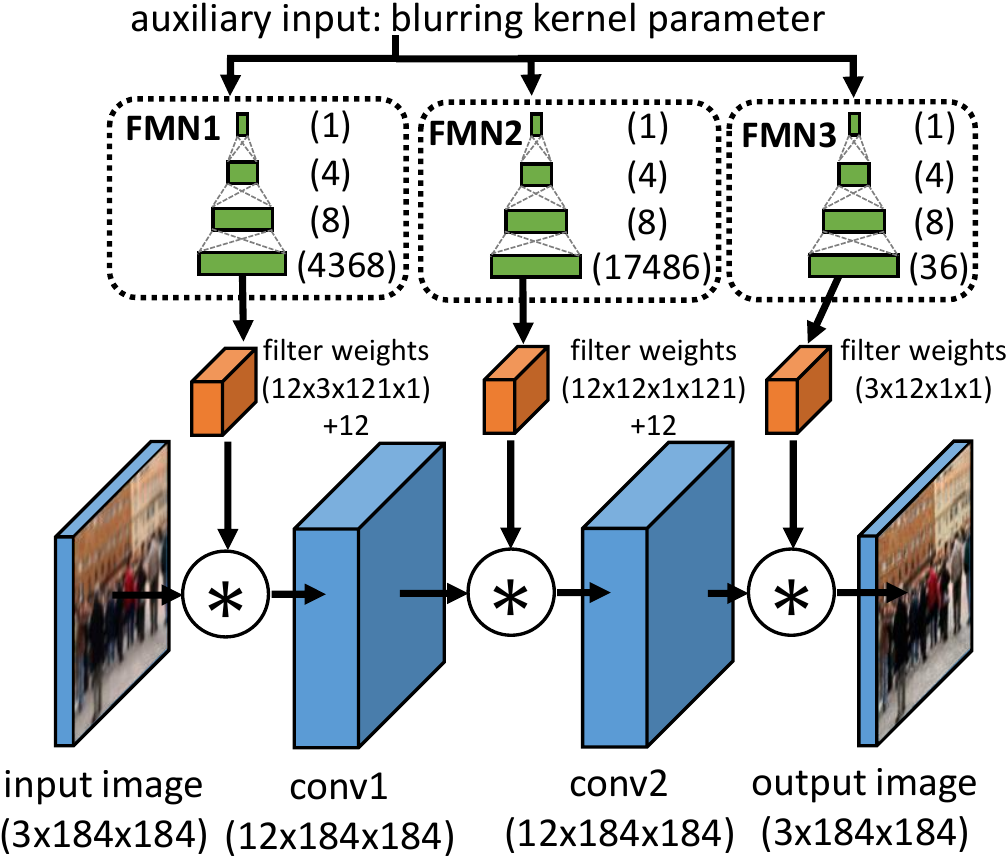}
\caption{ACNN for image deconvolution. The auxiliary input is the radius $r$ of the disk blurring kernel}
\label{fig:ACNNdeblur}
\end{figure}

Our ACNN for image deconvolution is based on the deconvolution CNN proposed in~\cite{Xu2014} (See Fig.~\ref{fig:ACNNdeblur}).
The input is the blurred image and the output target is the original image.
The ACNN consists of two long 1D adaptive convolutional layers: twelve 121$\times$1 vertical 1D filters, followed by twelve 1$\times$121 horizontal 1D filters. The result is passed through a 1$\times$1 adaptive convolutional layer to fuse all the feature maps.
We use leaky ReLU activations~\cite{Maas2013} for the first two convolutional layers, and sigmoid activation for the last layer to produce a bounded output as image.
Batch normalization layer~\cite{Ioffe2015} is used after the convolution layers.

\par
The number of filters in the standard CNN is 38 \cite{Xu2014}.
In order to balance the parameters for a fair comparison, our adaptive CNN uses 12 filter channels.
The dataset used for this task is Flickr8k~\cite{Hodosh2015,Rashtchian2010}. We randomly select 5000 images for training, 1400 images for validation and another 1600 images for testing.
The images were blurred uniformly using a disk kernel, and then corrupted with additive Gaussian noise (AWG) and JPEG compression as in~\cite{Xu2014}.
We train the models with images blurred with different sets of kernel radii $r\in\{3,5,7,9,11\}$.
The test set consists of images blurred with  $r\in\{3,5,7,9,11\}$.
The evaluation is based on the peak signal-to-noise ratio (PSNR) between the deconvolved image and the original image, relative to the PSNR of the blurred image.

\begin{table*}[tbh]
\small
\centering
\begin{tabular}{|c|c|ccccc|c|cc|}
  \hline
  Arch-filters  	& training set $r$ 	   & $r$=3   & $r$=5   & $r$=7   & $r$=9   & $r$=11  & all   & seen  $r$	& unseen $r$ \\
  \hline
  blurred image		& ---		   & 23.42   & 21.90   & 20.96   & 20.28   & 19.74   & 21.26 & ---  	& ---    \\
  CNN \cite{Xu2014} & \{7\}  	   & -8.20   &	-2.10   & +1.77	  & +0.21   & -0.73	  & -1.81 & +1.77   & -2.71  \\
  \hline
  CNN \cite{Xu2014} & \{3, 7, 11\} &  +0.55  &-0.25    & +0.49   &	+0.69   & +0.56   & +0.41 & +0.54   & +0.22 \\
  ACNN (ours)		& \{3, 7, 11\} & \bf{+0.77} & \bf{+0.06} & \bf{+1.17} & \bf{+0.94} & \bf{+1.28} & \bf{+0.84} & \bf{+1.08} & \bf{+0.50} \\
  \hline
  CNN \cite{Xu2014} & \{3, 5, 7, 9, 11\}  & +0.28 &	+0.45 &	+0.62 &	+0.86 &	+0.59 &	+0.56 & +0.56 & ---\\
  ACNN (ours)       & \{3, 5, 7, 9, 11\}  & \bf{+0.71} & \bf{+0.92} & \bf{+1.00} & \bf{+1.28} & \bf{+1.22} & \bf{+1.03} & \bf{+1.03} & --- \\
  \hline
\end{tabular}
\caption{PSNRs for image deconvolution experiments. The PSNR for the blurred input image is in the first row, while the other rows are the change in PSNR relative to that of the blurred input image.}
\label{tab:deconv}
\end{table*}


\par
Experimental results are shown in Table~\ref{tab:deconv} using different sets of radii for the training set.
First, when trained on the full training set, ACNN almost doubles the increase in PSNR, compared to the CNN (+0.56dB vs. +1.03dB).
Next, we consider a reduced training set with radii $r\in\{3,7,11\}$, and ACNN again doubles the increase in PSNR (+41dB vs. +0.84dB).  The performance of CNN on the unseen radii $r\in\{5,9\}$ is not as good as ACNN, which demonstrates the capability of ACNN to interpolate along the filter manifold for unseen auxiliary inputs.

\par
\begin{figure}[tb]
\centering
\footnotesize
\begin{tabular}{@{}c@{\hspace{1.5mm}}c@{}}
  (a) Original (target) & (b) Blurred (input) PSNR=24.34 \\
  \includegraphics[width=0.48\linewidth]{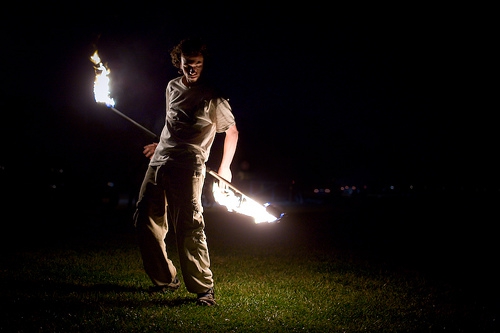} &
  \includegraphics[width=0.48\linewidth]{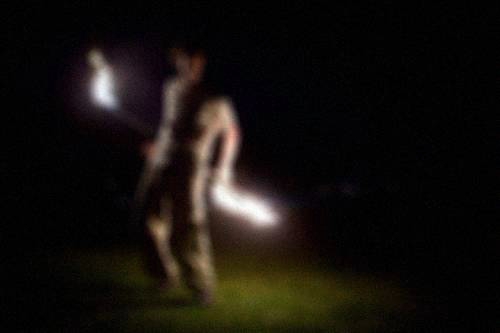} \\
  (c) CNN \cite{Xu2014} PSNR=25.30 &  (d) ACNN (ours) PSNR=26.04 \\
  \includegraphics[width=0.48\linewidth]{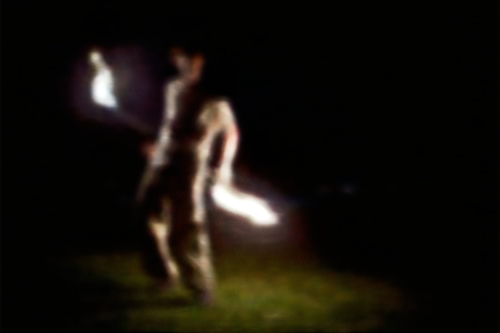} &
  \includegraphics[width=0.48\linewidth]{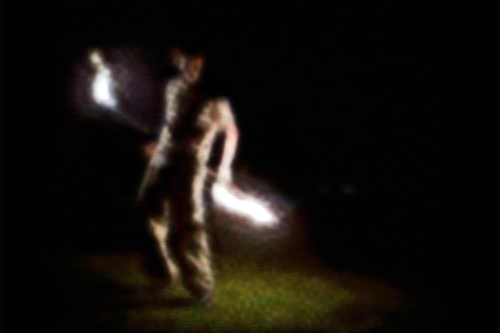} \\
\end{tabular}
\caption{Image deconvolution example: (a) original image; (b) blurred image -- the disk radius is 7; deconvolved images using (c) CNN-38 and (d) our ACNN.}
\label{fig:deconv_images}
\end{figure}
\begin{figure}[tb]
\centering
\small
\begin{tabular}{c}
  \includegraphics[width=0.98\linewidth]{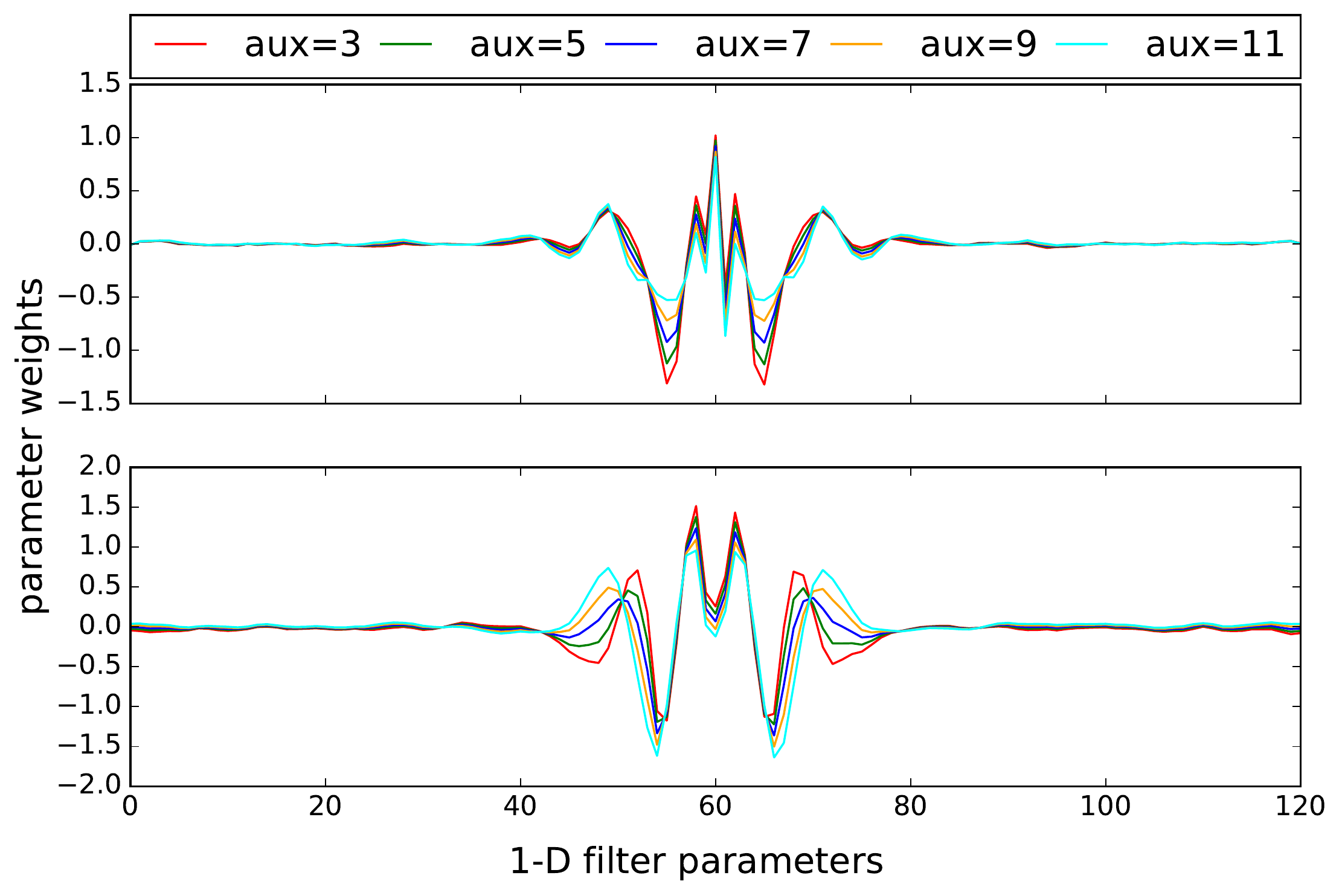}
\end{tabular}
\caption{Two examples of filter manifolds for image deconvolution. The y-axis is the filter weight, and x-axis is location. The auxiliary input is the disk kernel radius. Both the amplitude and the frequency can be adapted.}
\label{fig:deconv_filters}
\end{figure}

Fig.~\ref{fig:deconv_images} presents a visual comparison of deblurred images.
Deconvolved images using CNN are overly-smoothed since it treats images blurred by all the kernels uniformly.
In contrast, the ACNN result has more details and higher PSNR.
Example learned filters with respect to the auxiliary input (radius of the disk kernel) are shown in Fig.~\ref{fig:deconv_filters}.

\section{Conclusion}

In this paper, we propose an adaptive convolutional neural network (ACNN), which employs the available side information as an auxiliary input to generate the convolutional filter weights.
The ACNN can disentangle variations related to the side information, and extract features related to the current context.  We apply ACNN to single-scene crowd counting, using the perspective weight as the side information, and cross-scene crowd counting, using the camera angle and height as side information, and show that ACNN has improved learning capacity compared to a plain CNN with a similar number of parameters.
We also apply ACNN to image deconvolution, and show that it is possible to train one network to handle different kernel parameters.
Future work will consider using ACNN for other computer vision tasks, e.g., stereo matching where the disparity is the auxiliary input.

{\small
\bibliographystyle{IEEEtran}
\bibliography{ref_dl_fg_counting.bib}
}

\end{document}